\documentclass[letterpaper]{article}
\usepackage{aaai}
\usepackage{times}
\usepackage{helvet}
\usepackage{courier}
\usepackage{graphicx}
\graphicspath{ {./images/} }
\begin{document}
%
\title{DeepAg: Deep Learning Approach for Measuring the Effects of Outlier Events on Agricultural Production and Policy}

\author{Sai Gurrapu\textsuperscript{1}, Feras A. Batarseh\textsuperscript{1}, Pei Wang\textsuperscript{1}, Md Nazmul Kabir Sikder\textsuperscript{1}, \\  
\Large{\bf{Nitish Gorentala\textsuperscript{1}, Gopinath Munisamy\textsuperscript{2}}}\\

\textsuperscript{1} College of Engineering, Virginia Polytechnic Institute and State University (Virginia Tech), Arlington, VA, USA\\
\textsuperscript{2} Department of Agricultural and Applied Economics, University of Georgia, Athens, GA, USA \\
\{saig, batarseh, pwang1, nazmulkabir, nitishg\}@vt.edu, m.gopinath@uga.edu\\
}

\maketitle
\begin{abstract}
\begin{quote}
Quantitative metrics that measure the global economy’s equilibrium have strong and interdependent relationships with the agricultural supply chain and international trade flows. Sudden shocks in these processes caused by outlier events such as trade wars, pandemics, or weather can have complex effects on the global economy. In this paper, we propose a novel framework, namely: DeepAg, that employs econometrics and measures the effects of outlier events detection using Deep Learning (DL) to determine relationships between commonplace financial indices (such as the DowJones), and the production values of agricultural commodities (such as Cheese and Milk). We employed a DL technique called Long Short-Term Memory (LSTM) networks successfully to predict commodity production with high accuracy and also present five popular models (regression and boosting) as baselines to measure the effects of outlier events. The results indicate that DeepAg with outliers' considerations (using Isolation Forests) outperforms baseline models, as well as the same model \textit{without} outliers detection. Outlier events make a considerable impact when predicting commodity production with respect to financial indices. Moreover, we present the implications of DeepAg on public policy, provide insights for policymakers and farmers, and for operational decisions in the agricultural ecosystem. Data are collected, models developed, and the results are recorded and presented. 
\end{quote}
\end{abstract}

\section{Introduction and Motivation}
\noindent According to the United Nations, the world’s population is expected to increase by two billion persons in the next 30 years, from 7.7 billion (current) to 9.7 billion in 2050 and could peak at nearly 11 billion around 2100. To feed this growing population, a similar increase in food production must also be achieved \cite{KAMILARIS25}. Several challenges exist in agriculture -with declining productivity of resources such as land, the environmental footprint of production practices, and the ensuing need for sustainability– limiting human abilities to scale up production to meet the global demand. The integration of technology into the agricultural ecosystem is considered to be an important pathway for not only providing adequate nutrition to the world but also ensuring sustainability of the resources for the benefit of future generations \cite{Liakos1}. For instance, precision farming where producers optimize their complex multivariate farming practices by continuously monitoring, measuring and analyzing several variables such as weather, soil, and crop type, enables precise targeting and care for each specific agricultural commodity at scale that was not possible in the 20th century \cite{Gebbers26}. However, with agriculture highly susceptible to outlier events, e.g., floods, drought, and trade wars, predicting the future with an accounting for possible outlier events remains a major challenge \cite{Gopinath6}. Big data analytics in precision farming demonstrated the importance of recognizing and extracting insights and trends from historical agricultural data to better guide commodity production decisions and policy making based on context \cite{Storm18}. As the amount of data generated in the agricultural ecosystem continues to increase, Machine Learning (ML) is being used to provide accurate predictive insights and guidance on operational decisions with real-time data \cite{Wolfert27}.

ML allows the machine to learn from the available data without being explicitly programmed, thus revealing more insights than what is normally possible through traditional data analytics. DL extends classical ML by adding more complexity into the models with a large learning capacity as it has a strong advantage in feature learning. This enables DL models to be flexible and highly adaptable for a wide range of complex tasks. That notion allows it to excel at classification and prediction problems in many domains \cite{KAMILARIS25}. The application of DL in agriculture is relatively recent and can be a promising technique considering the impact and potential it has demonstrated in other domains \cite{KAMILARIS25}. Most studies and applications of precision farming today are localized to the farm environment without much consideration of the impact of external variables, specifically outlier events \cite{GurrapuPoster}. The hypothesis tested in our work is twofold: (1) Whether DL will perform better than traditional statistical (such as regression) techniques for predicting multivariate relationships in farming? (2) Outlier events, such as Economic, Financial, Weather, or Political events influence precision farming practices -regardless if they are positive or negative- and therefore need to be accounted for as an important part of the life-cycle process, can a combination of isolation forests and LSTM accomplish that with high accuracy? During the COVID-19 pandemic for instance, many farmers and producers were struggling with the forecasts provided to them using traditional econometrics because the models used to create such predictions don't account for outlier events. 

Accordingly, in this manuscript, we propose an approach called DeepAg that employs DL to measure the effect of outlier events on agricultural production and employ it to predict the future production patterns in the presence of outliers. For this purpose, data on five commonplace financial indices and agricultural commodities production, along with that on outliers’ events over the past two decades are assembled on a monthly scale. A combination of isolation forests and LSTM model is employed to train and test for classification. The input features include the commodities production (data collected from United States Department of Agriculture - USDA) and their respective highest \textit{causation} and \textit{correlation} with market’s financial indices. We present two forecasts for each commodity’s production, one with the presence of an outlier event, and one without. To demonstrate the effectiveness of our proposed approach and the importance of outliers' detection, we also present several baseline models using standard ML techniques and compare that with an LSTM model that doesn’t consider outliers. The results indicate that the LSTM model with isolation forests (i.e. outlier detection) outperforms all others and can be used to determine the relationship between commodity production and financial indices with a high accuracy, in the presence of outliers.

The rest of this paper is organized as follows: The Related Work section presents relevant literature on the developments of ML and DL for precision farming; the Experiment section describes the datasets used for the experiment and overall experimental setup; the Methodology section introduces our approach and outlines its evaluation; the Results section discusses the outcomes of our experiment; and finally, the Conclusions section evaluates our approach and provides insights on precision farming and the importance of outlier events detection for farmers and policy makers.

\section{Related Work}
As agricultural ecosystems adopt technology to improve their farming practices, the data collected in the background is increasingly valuable. In \cite{Liakos1}, a comprehensive review was conducted on ML applications for agricultural production systems. They demonstrated examples of certain precision farming practices such as crop and soil management, disease detection, livestock management, and water usage amongst others that can be improved using ML. A Support Vector Machines (SVM) based methodology was presented by \cite{MORALES3} for the early detection of problems in egg production. The experiment forecasts egg production for up to three days and sends an alert if the production curve displays any anomalies. The results demonstrate that a poultry management system with production forecasting would prove to be useful to assist producers with preventative measures before a problem occurs. Another approach using SVM was shown in \cite{ALONSO4} to predict the weight trajectory of livestock given the past evolution of the herd. Additionally, using advanced hardware sensing techniques and artificial neural networks, \cite{PANTAZI5} demonstrated an architecture to predict wheat yield production with a high accuracy of 91\%.

Agricultural data has also been shown to be useful outside of the farm environment. For example, \cite{Gopinath6} and \cite{gopinath7} employed deep-learning ML techniques (unsupervised and supervised) to predict trade patterns of seven major agricultural commodities and indicated that unsupervised ML approaches with neural networks provides better prediction fits over the long term. A method in \cite{Monken8} was proposed to measure causal scenarios in trade during outlier events using network-based models, specifically Graph Neural Networks (GNNs) were used to predict outliers effectively and to provide relevant domain explainability. In \cite{Batarseh9}, Association Rules (AR) analysis was employed to identify imports and exports associations (\textit{if a then b}) with the trade flows and used Ensemble Machine Learning (EML) methods for agricultural trade predictions. In \cite{Storm18}, the use of ML for econometric practices was presented and demonstrates the challenges of such simulation models and shortcomings when used for quantitative economic analysis. A fast unsupervised algorithm called Isolation Forest was proposed by \cite{Liu10} for detecting anomalies in continuous data (of all domains). Accordingly, no method is found that could be applied to the production of all commodities considering multiple forms of outliers, our study aims to address that gap. 

\section{Experimental Work}
This section describes the datasets used for the experiment, the indices, outlier detection methods, and the overall pipeline.
\subsection{Financial Indices Dataset}
A collection of financial indices data are obtained from Yahoo Finance (https://finance.yahoo.com/). Daily market close data are collected for years 2000-2019, the following indices are considered: Gold, Crude Oil, Dow Jones Industrial Average (DOW), S\&P 500, and Volatility Index (VIX). Significance of the five financial indices is as follows:
\begin{itemize}
\item \textit{Gold}: Price of one ounce of gold \cite{Toraman14}.
\item\textit{ Crude Oil}: Price of a barrel of benchmark crude oil \cite{Toraman14}.
\item \textit{DOW}: Index of thirty most prominent companies listed on US stock exchanges \cite{Weiwei15}.
\item\textit{ S\&P} 500: Index of 500 of the largest companies listed on US stock exchanges \cite{Weiwei15}.
\item \textit{VIX}: Measure of the market’s expected volatility expectation based on the S\&P 500 index \cite{Hirsa16}.
\end{itemize}

\subsection{Commodities Dataset}
The data on the commodities was acquired from USDA's National Agricultural Statistics Service (NASS) QuickStats database. The commodities dataset was cleaned and segmented based on the commodity type. The list of commodities is as follows: \textit{Beef, Butter, Cheese, Chickens, Ducks, Eggs, Ice Cream, Lamb and Mutton, Milk, Other poultry, Pork, Sherbet, Turkeys, Veal, and Water Ices}. The available frequency of the commodities is the first day of every month, however, the financial indices frequency is for every weekday. This created a conflict when merging the two datasets because not every first day of the month is a weekday. In this case, we supplemented the data by labeling the most recent weekday as the first day of the month. 

\subsection{Outlier Events}
Outlier detection can be defined as a rare event identification, an observation that differs significantly from the rest of the data points \cite{Liu10}. Anomalies or outlier events in our experiment typically indicate a major disturbances related to the Economy (eg. trade war), Financial (eg. recession), Weather (eg. natural disaster), or Political (eg. government instability). We use anomaly detection algorithm to find data points that show abnormal behavior from the rest of the data points in financial indices. A proper understanding of the outlier events will help us better predict the relationship between financial indices and commodity production. We used an unsupervised algorithm, \textit{Isolation Forest}, for detecting anomalies. This algorithm works best if the outlier points are easily isolable and contamination rates of abnormal points are low. The overall pipeline of DeepAg is presented in Figure 1.

\section{Methodology}
\begin{figure}[h]
\centering
\includegraphics[width=\linewidth]{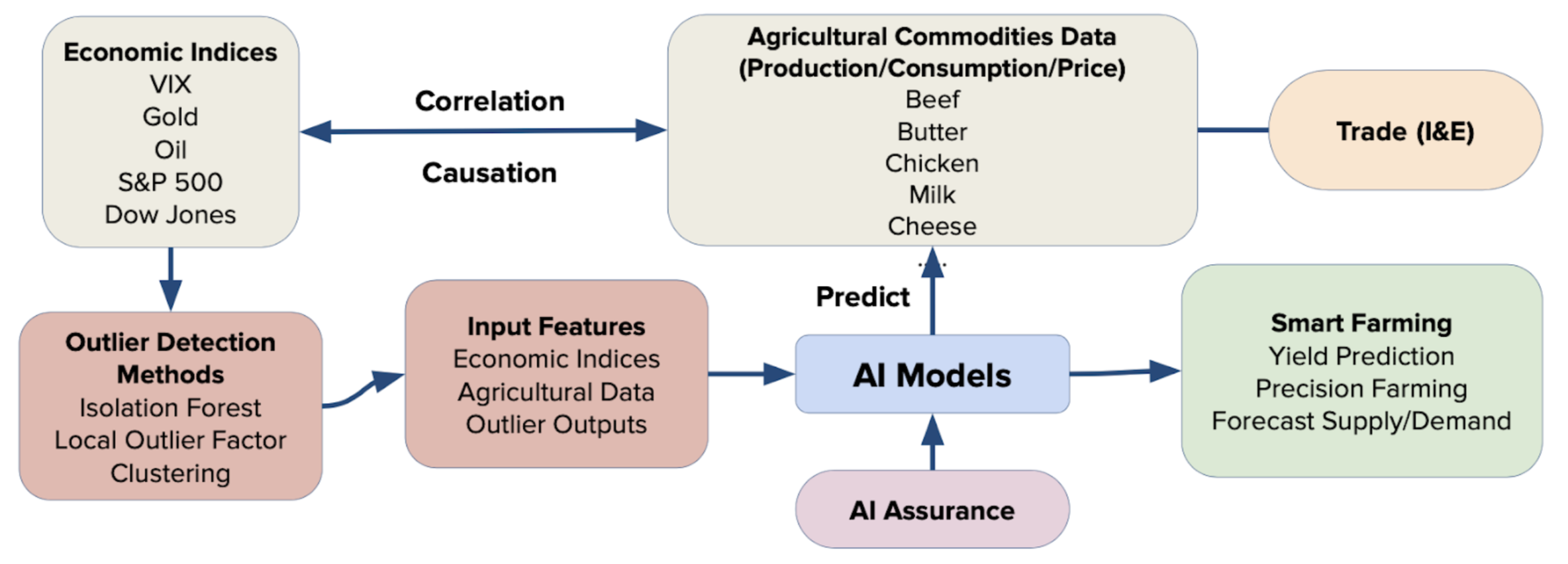}
\caption{The DeepAg methodology}
\label{fig:methodology}
\end{figure}

\subsection{Data Preprocessing}
All data are pre-processed before being fed into the baselines or LSTM models. To minimize bias, we employed data transformation techniques to normalize each of the input features using MinMaxScaler as represented in equation \ref{eq:minmax}.

\begin{equation}
x_{scaled}=\frac{x - x_{min}}{x_{max} - x_{min}}
\label{eq:minmax}
\end{equation}

\begin{figure}[h]
\centering
\includegraphics[width=\linewidth]{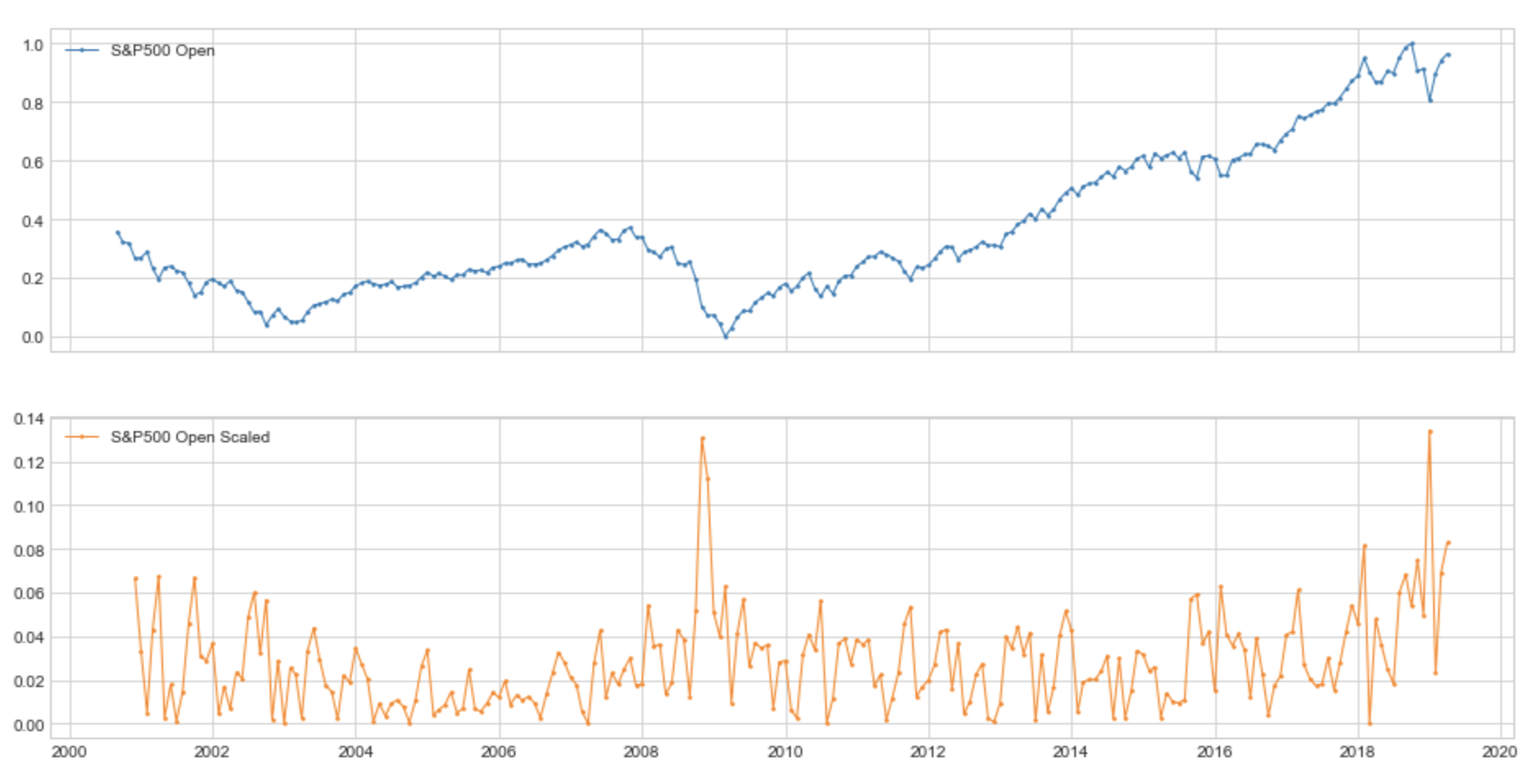}
\caption{Original S\&P 500 data from 2000-2019 (top) and scaled data (bottom)}
\label{fig:outliersScaled}
\end{figure}

Formula 1 normalizes the values of the financial indices dataset, commodities dataset, and the outlier events dataset into a range of \textit{0-1}. To prepare the datasets for anamoly detection, we then used \textit{DoubleRollingAggregate} from the \textit{ATDK Python} library to track the statistical behavior in a time series dataset. The DoubleRollingAggregate transformer rolls two sliding windows side-by-side along with a time series, aggregates using statistical mean, and tracks the difference of the aggregated metrics between the two windows. Figure \ref{fig:outliersScaled} shows the changes (normalization) to the indices after applying \textit{DoubleRollingAggregate} Transformer.

\subsection{Outlier Detection}
The attribute of Isolation Forest is that there is a tendency to separate outlier data points from normal data points, as the algorithm randomly selects an attribute and splits values between the minimum and maximum of that attribute. The outlier detection in our work is based on having the economic indices as an input to the Isolation Forest algorithms as shown in Figure \ref{fig:methodology}. Afterwards, it recursively creates random partitions to find outlier points. Random partitioning creates a short path for outliers. Therefore, a sample is more likely to be an outlier if a forest of random trees collectively produces shorter path lengths for particular samples. 



\subsubsection{Isolation Forest}
The algorithm takes several hyperparameters as input and among them the most important is \textit{contamination rate}. It decides the percentage of outliers which can be approximately predicted out of the total data points. We determine the contamination rate using a popular statistical measure called Interquartile Range (IQR). IQR describes the middle \textit{50\%} of the data distribution. Quartiles slice any Gaussian distribution into four equal groups of \textit{25\%}. Calculating IQR describes the middle half of the data in the distribution set. We are considering these middle data segments as normal data points since outliers are usually at the tails of a Gaussian distribution. The IQR can be represented as:

\begin{equation}
\textrm{$Interquartile\: Range$} = Q3 - Q1
\end{equation}

Where Q1 is the first quartile or 25th percentile and Q3 is the third quartile or 75th percentile. Tables \ref{tab:dailyIQR} and \ref{tab:monthlyIQR} provide the daily and monthly contamination rate of all financial indices using the IQR method. 
The Isolation Forest algorithm identifies an estimation of the anomaly score for a given instance \textit{x} using this formula:

\begin{equation}
s(x, m) = 2 ^{-E (h(x))/c(m)}
\end{equation}
Where \textit{E} is the average of \textit{h} trees and \textit{c(m)} is the average path length of unsuccessful binary searches.

\begin{table}[h]
\centering
\scriptsize{
\begin{tabular}{|l|l|}
\hline
\multicolumn{2}{|l|}{\textbf{Daily Data Contamination (Using IQR)}} \\ \hline
\textbf{Financial Index}        & \textbf{Contamination Rate}       \\ \hline
VIX                             & 6.559                             \\ \hline
Gold                            & 5.382                             \\ \hline
S\&P 500                        & 6.008                             \\ \hline
DOW                             & 6.125                             \\ \hline
Crude Oil                       & 3.953                             \\ \hline
\end{tabular}}
\caption{\label{tab:dailyIQR}Daily IQR contamination for financial indices}
\end{table}

\begin{table}[h]
\centering
\scriptsize{
\begin{tabular}{|l|l|}
\hline
\multicolumn{2}{|l|}{\textbf{Monthly Data Contamination (Using IQR)}} \\ \hline
\textbf{Financial Index}         & \textbf{Contamination Rate}        \\ \hline
VIX                              & 6.250                              \\ \hline
Gold                             & 2.232                              \\ \hline
S\&P 500                         & 2.232                              \\ \hline
DOW                              & 2.232                              \\ \hline
Crude Oil                        & 6.250                              \\ \hline
\end{tabular}}
\caption{\label{tab:monthlyIQR}Monthly IQR contamination for financial indices}
\end{table}

\subsubsection{Algorithm Accuracy}
\begin{figure}[h]
\centering
\includegraphics[width=\linewidth]{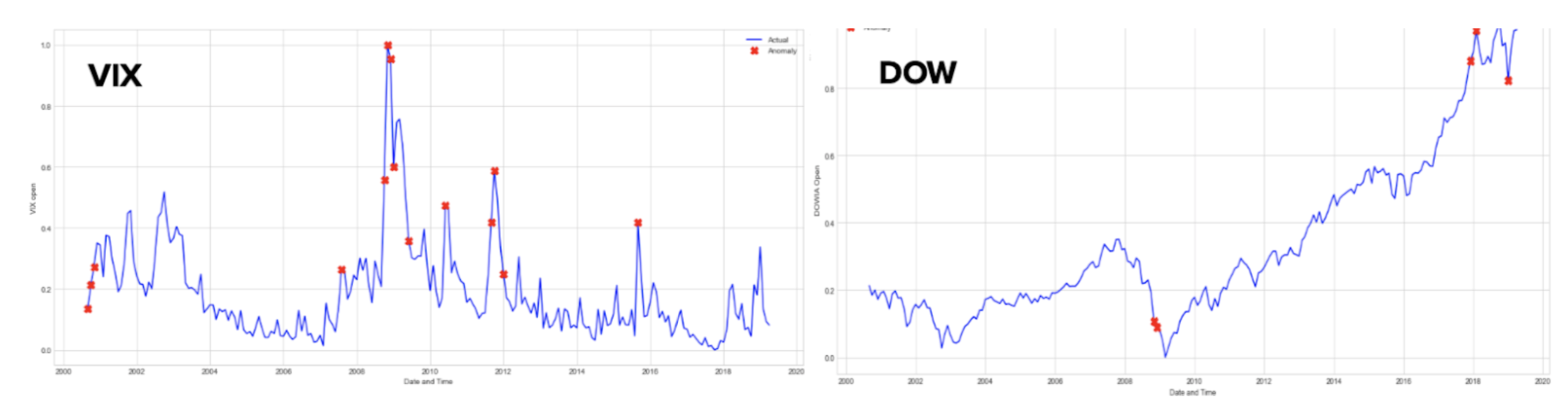}
\caption{Graph of anomalies present in VIX and DOW}
\label{fig:vixDowAnomaly}
\end{figure}

After determining the anomalies, we plotted the data as shown by Figure \ref{fig:vixDowAnomaly}. The red markers on the graphs indicate the outlier events found by the isolation forest algorithm. To ensure high accuracy, we manually evaluated the anomalies by cross-checking them with historical events that have occurred during that time. The algorithm proved to be accurate in determining and isolating outlier events as it was able to identify all events such as trade war, COVID-19, hurricanes, and others. 

\subsection{Causation and Correlation}
\begin{figure}[h]
\centering
\includegraphics[width=\linewidth]{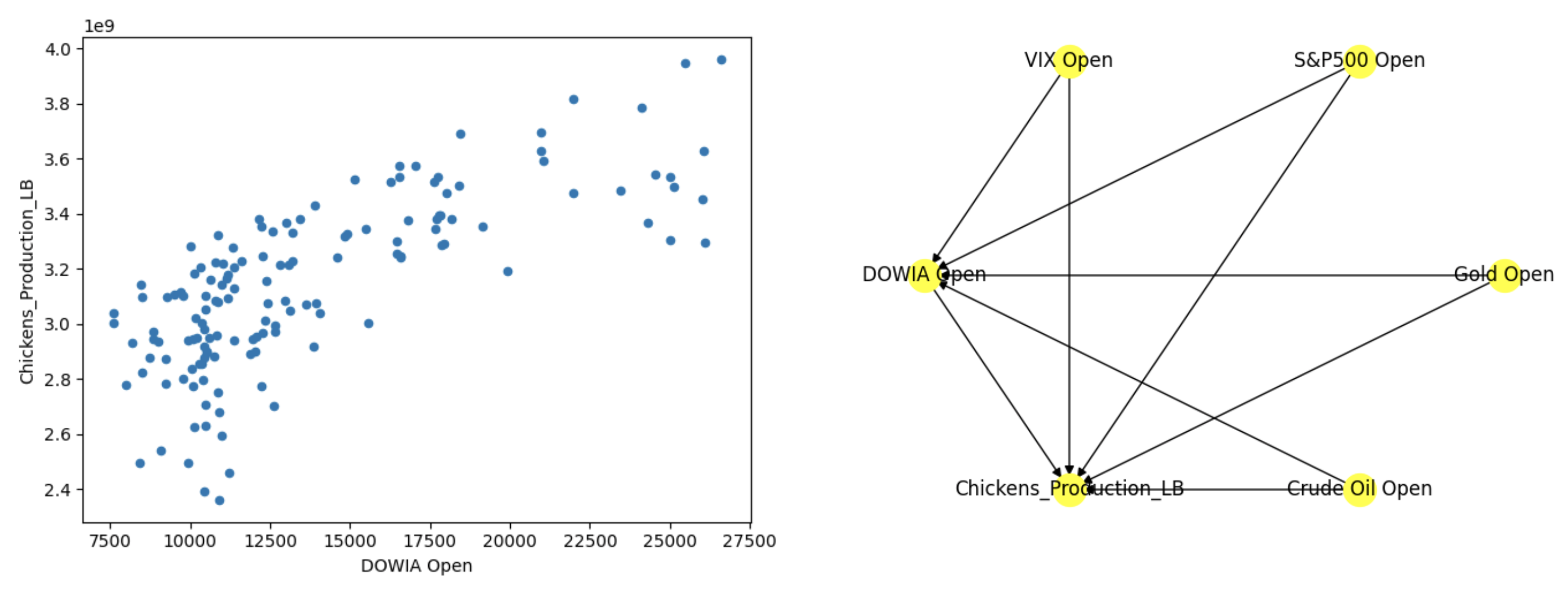}
\caption{Chicken production and DOW correlation (Left); chicken production and DOW causation (right)}
\label{fig:chickenCorr}
\end{figure}

\begin{figure}[h]
\centering
\includegraphics[width=\linewidth]{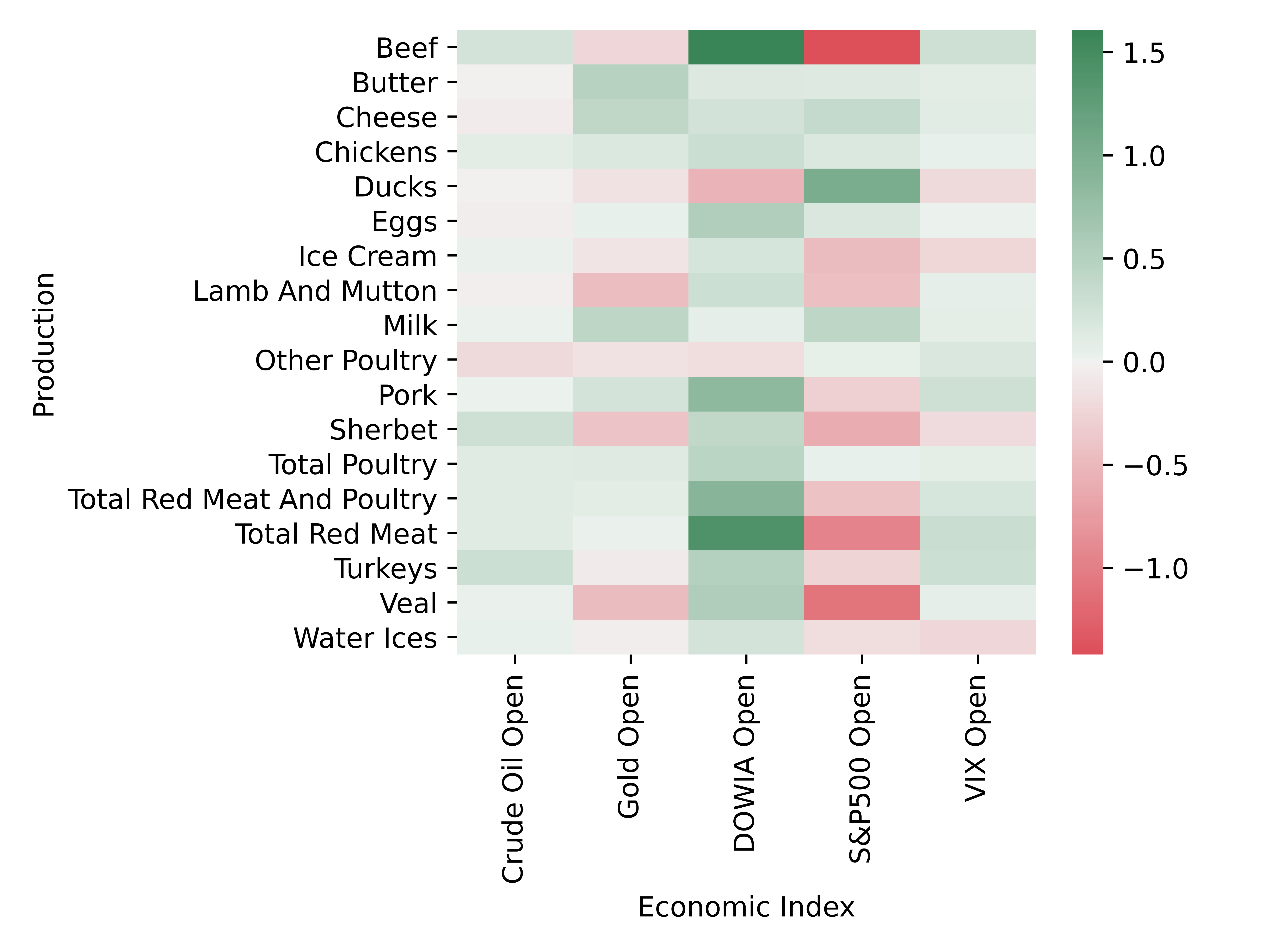}
\caption{Causation matrix of commodities and indices}
\label{fig:causationMatrix}
\end{figure}

The \textit{DoWhy Python} library is used for causal inference and for measuring causality. Measuring causality enabled us to use the financial index that had the greatest causal impact on each agricultural commodity (for instance, commodity: \textit{Milk} had a high causation score with Index: \textit{Gold}). While correlation (Figure \ref{fig:chickenCorr}) shows whether two measures are following similar trends in a dataset, causation (Figure \ref{fig:chickenCorr}) can be useful to determine what is truly causing another dataset to change. It's evident from the results that the causation matrix in Figure \ref{fig:causationMatrix} \textit{DOW} and \textit{meat} commodities generally have a strong causal relationship with green color indicating high causality, and red indicating low. In the Table \ref{tab:CorrelationTable} of correlation results, VIX indicates the lowest correlation with the commodities compared to the other indices. Moreover, essential commodities such as Butter, Cheese, and Chickens seem to be strongly correlated with important economic indicators such as DOW and S\&P 500.

\begin{table}[h!]
\renewcommand{\tabcolsep}{6.5pt}
\scriptsize{
\begin{tabular}{|p{0.24\linewidth}|r|r|r|r|r|}
\hline
\textbf{Commodities}                & \textbf{GOLD}  & \textbf{DOW} & \textbf{S\&P 500} & \textbf{OIL} & \multicolumn{1}{l|}{\textbf{VIX}} \\ \hline
\textbf{Beef}                       & 0.096 & -0.060       & -0.030           & -0.043       & 0.043                             \\ \hline
\textbf{Butter}                     & -0.097         & 0.721        & 0.571            & 0.529        & 0.403                             \\ \hline
\textbf{Cheese}                     & -0.276         & 0.801        & 0.863            & 0.825        & 0.371                             \\ \hline
\textbf{Chickens}                   & -0.307         & 0.676        & 0.748            & 0.709        & 0.389                             \\ \hline
\textbf{Ducks}                      & -0.427         & 0.110        & 0.556            & 0.564        & -0.088                            \\ \hline
\textbf{Eggs}                       & -0.502         & 0.147        & 0.869            & 0.863        & -0.406                            \\ \hline
\textbf{Ice Cream}                  & -0.054         & -0.277       & -0.256           & -0.230       & -0.108                            \\ \hline
\textbf{Lamb \& Mutton}            & 0.182          & -0.771       & -0.558           & -0.505       & -0.490                            \\ \hline
\textbf{Milk}                       & -0.261         & 0.803        & 0.781            & 0.744        & 0.427                             \\ \hline
\textbf{Other poultry}              & 0.183          & -0.462       & -0.322           & -0.283       & -0.400                            \\ \hline
\textbf{Pork}                       & -0.089         & 0.678        & 0.701            & 0.652        & 0.329                             \\ \hline
\textbf{Sherbet}                    & -0.005         & -0.418       & -0.395           & -0.377       & -0.064                            \\ \hline
\textbf{Turkeys}                    & 0.046          & 0.289        & 0.205            & 0.175        & 0.287                             \\ \hline
\textbf{Veal}                       & 0.319          & -0.791       & -0.808           & -0.772       & -0.414                            \\ \hline
\textbf{Water Ices}                 & -0.197         & -0.019       & 0.054            & 0.070        & 0.010                             \\ \hline
\end{tabular}}
\caption{\label{tab:CorrelationTable}Correlation of commodities and indices}
\end{table}

\subsection{Artificial Intelligence Models for DeepAg}
\subsubsection{Baseline Models}
To ensure that our proposed DeepAg LSTM approach works effectively, and to determine if it outperforms existing approaches, baseline models are created for evaluation purposes. We applied five popular boosting and regression models as baselines: \textit{Linear Regression, Linear Regression with Polynomial Features, Regression Tree, Random Forest Regressor}, and \textit{XGBoost Regressor} to predict commodities' production. The input for the regression model includes the highest causation and correlation financial indices and the historical commodity production data. Every commodity is paired with an index (ones with the \textit{highest} correlation and causation score per commodity).

\subsubsection{DeepAg LSTM Approach}
The architecture of Recurrent Neural Networks(RNN) has allowed for more accurate modeling of time-series processes where references to information from the past persist \cite{lipton12}. The Long Short-Term Memory (LSTM) model improves upon the RNN architecture to overcome the vanishing gradient problem and enables sequence learning with high accuracy \cite{Hochreiter13}. The LSTM model is an optimal choice for our experiment considering its state-of-the-art performance in sequence learning and our dataset is heavily time-series based. The process of preserving information and carrying it forward occurs in the LSTM memory cell and it is composed of Input Gate ($i_{t}$), Forget Gate ($f_{t}$), and Output Gate ($o_{t}$). The mathematical representation of an LSTM memory cell can be represented as follows (in formulas 3, 4, and 5):

\begin{equation}
i_{t} = \sigma(w_{i}[h_{t-1},x_{t}] + b_{i})
\end{equation}

\begin{equation}
f_{t} = \sigma(w_{f}[h_{t-1},x_{t}] + b_{f})
\end{equation}

\begin{equation}
o_{t} = \sigma(w_{o}[h_{t-1},x_{t}] + b_{o})
\end{equation}

Where $\sigma$ represents sigmoid function, $w_x$ is the weight for the respective gate$(x)$ neurons, $h_{t-1}$ is the output of the previous LSTM block at timestamp $t-1$, $x_t$ is the input at current timestamp, and $b_x$ is the biases for the respective gates$(x)$ \cite{Hochreiter13}. Once the cell state is filtered, it is passed through an activation function, specifically the sigmoid function, and it predicts the output, $h_t$,  of the LSTM unit at timestamp $t$. The output, $h_{t}$, is passed through a softmax layer to obtain the predicted output $y_t$. This process is represented as follows:

\begin{equation}
\tilde{c}_{t} = tanh(w_{c}[h_{t-1},x_{t}] + b_{c})
\end{equation}

\begin{equation}
c_{t} = f_t * c_{t-1} + i_t * \tilde{c}_{t}
\end{equation}

\begin{equation}
h_{t} = o_t * tanh(c^t)
\end{equation}

Where $c_t$ is the cell state memory at timestamp $t$ and $\widetilde{c}_t$ represents candidate for cell state at timestamp $t$.

The objective of the LSTM model is to better predict the multivariate relationship between the commodity production, and financial indices \textit{with} and \textit{without} considering outlier events. In order to measure the effects of outliers on the prediction, the final LSTM model is developed where one prediction is made with the outlier events as one of the input features (outputted by isolation forest) and another prediction without. In the case that the highest correlation and causation were equal, we performed the same procedure as we did for the baselines of selecting only one of the indices with a preference towards causation scores.

\section{Results}
\subsection{Baseline Models}
\begin{figure}
\centering
\includegraphics[width=\linewidth]{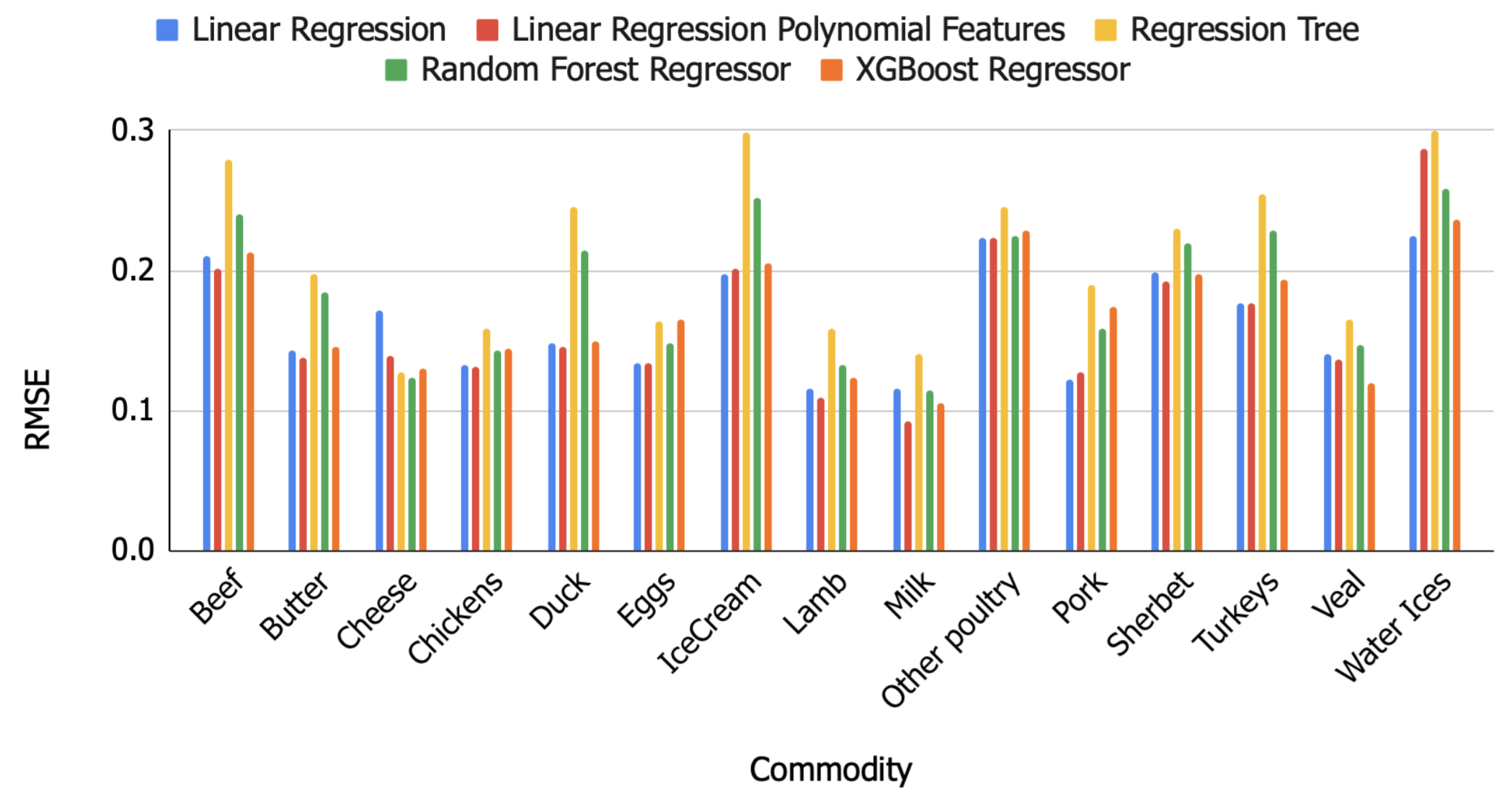}
\includegraphics[width=\linewidth]{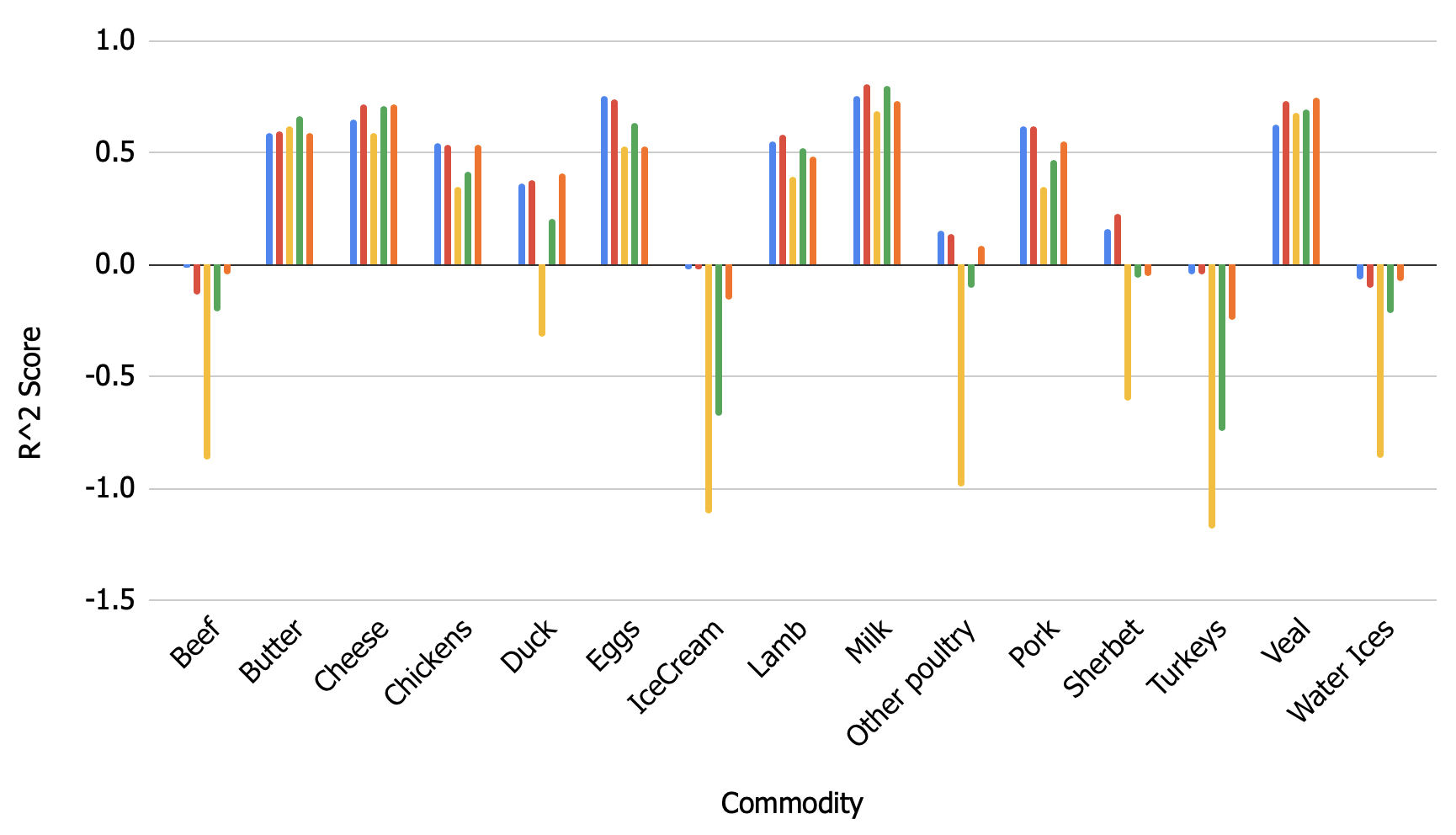}
\caption{\label{fig:BaselineModelsResults}Baseline models Root Mean Squared Error (\textit{RMSE}) and $R^2$ scores results}
\end{figure}

Our experiment with the baseline models included default hyperparameters with no feature engineering. The results rank the overall performance of the models as follows: Linear Regression with Polynomial Features, Linear Regression, XGBoost Regressor, Random Forest Regressor, and Regression Tree. \textit{Linear Regression with Polynomial Features} resulted with the most accurate and best-performing baseline model (as shown in Figure \ref{fig:BaselineModelsResults}).

We use the $R^2$ score to determine the accuracy of the regression models as calculated by equation \ref{eq:R2Eq} and \textit{RMSE} to evaluate the error as shown in equation \ref{eq:RMSEEq} given our prediction, $\bar{y}$, and the actual value, $y$. The commodity production values typically range in the billions/year depending on the commodity.

\begin{equation}
\label{eq:R2Eq}
R^2 = 1 - \frac{SS_{\textrm{$RES$}}}{SS_{\textrm{$TOT$}}} = 1 - \frac{\sum_i (y_i - \hat{y}_i)^2}{\sum_i (y_i - \bar{y}_i)^2}
\end{equation}

\begin{equation}
\label{eq:RMSEEq}
\textrm{$RMSE$} = \sqrt{\sum_{i = 1}^{n}{\frac{(\hat{y}_i - y_i)^2}{n}}}
\end{equation}

When comparing $R^2$ scores across models, the linear fit can be seen as unusually low for certain commodities such as Beef, Ice Cream, and Water Ices. This is most likely due to the variance in the data (i.e. variable production due to economic factors). In addition, USDA has more available historical production data for these commodities compared to others and this also may have been a factor in affecting the fit.

\begin{figure}[h]
\centering
\includegraphics[width=\linewidth]{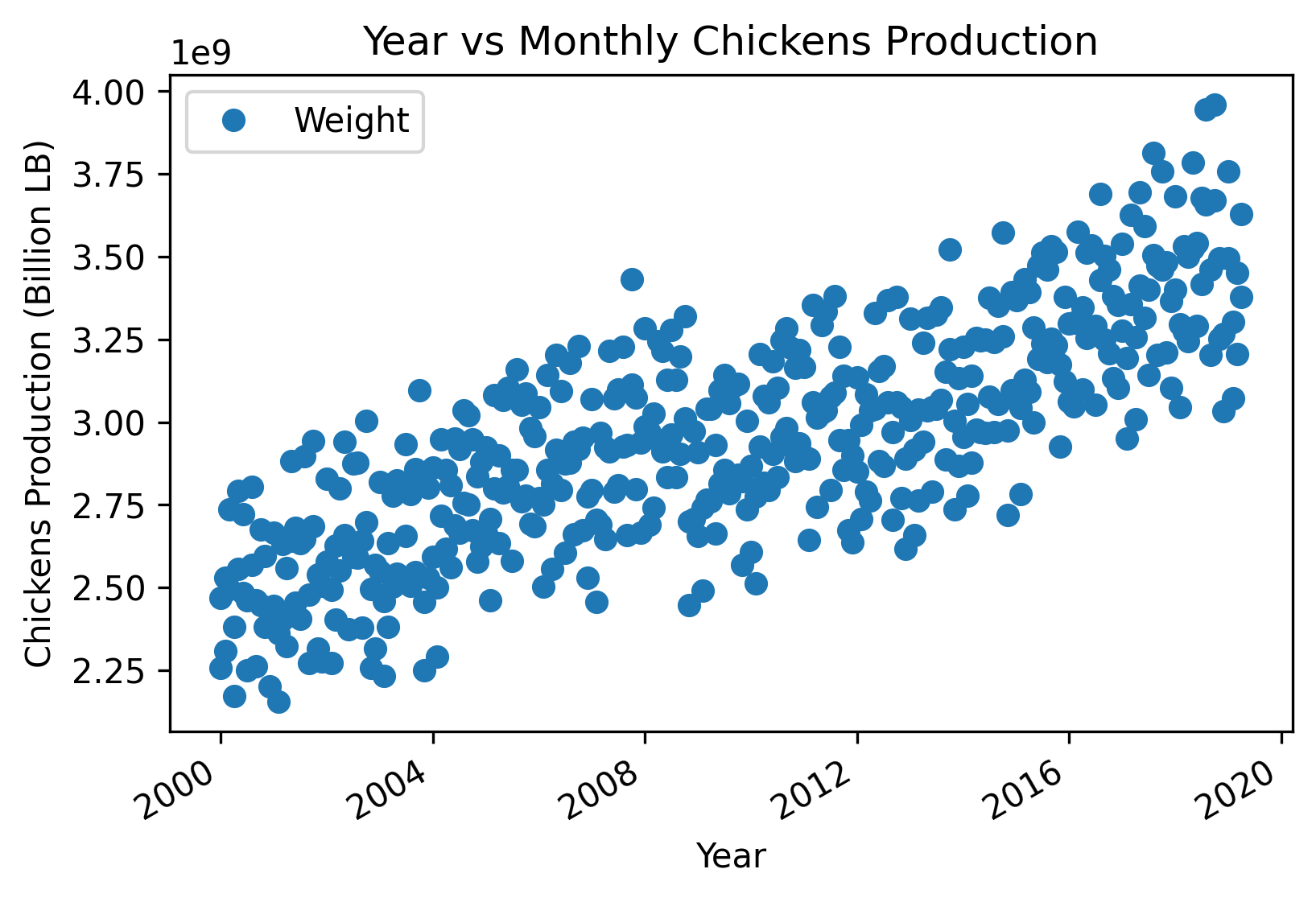}
\caption{Historical chickens production}
\label{fig:chickensProd}
\end{figure}

\begin{figure}[h]
\centering
\includegraphics[width=\linewidth]{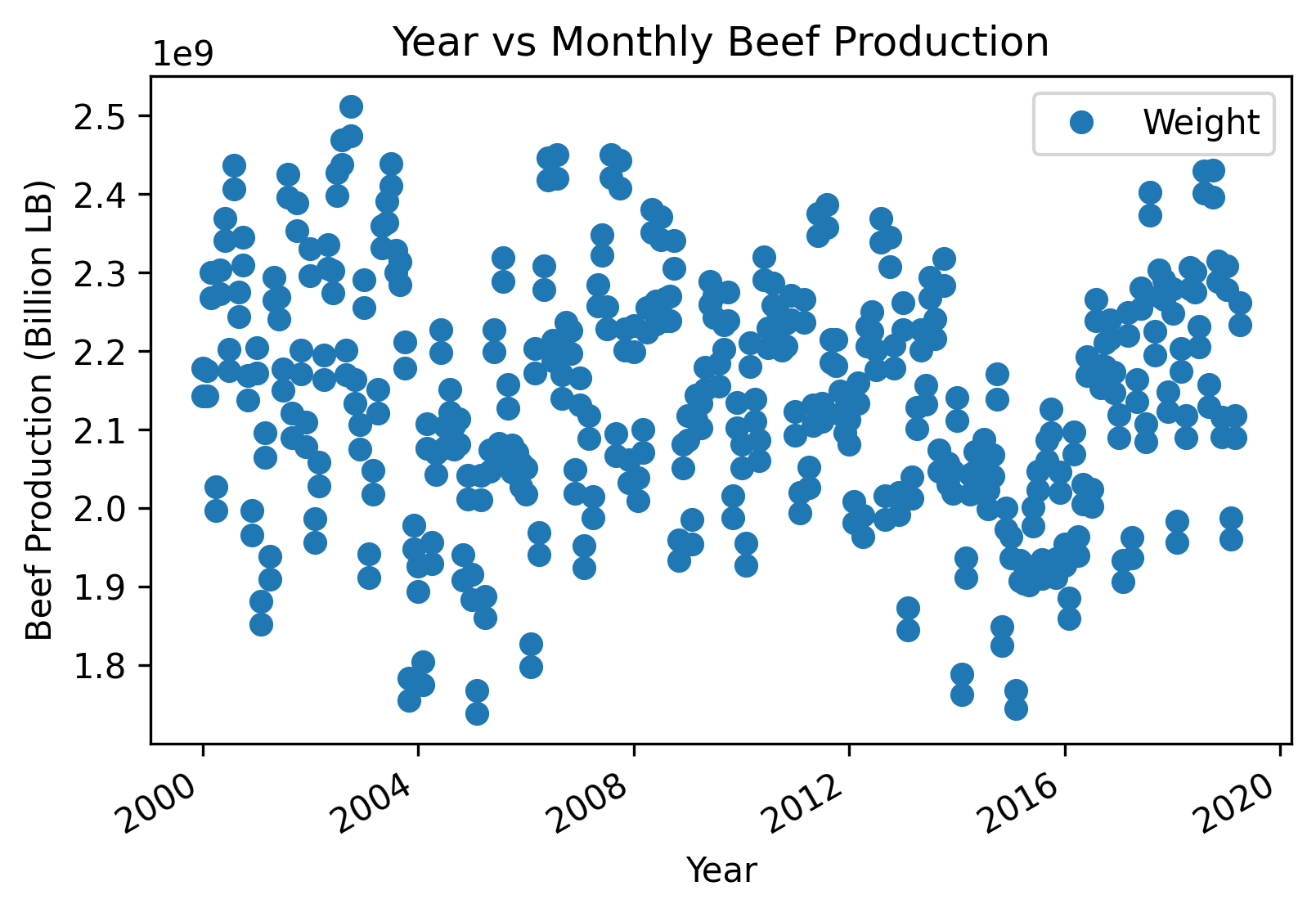}
\caption{Historical beef production}
\label{fig:beefProd}
\end{figure}

As shown in Figures \ref{fig:chickensProd} and \ref{fig:beefProd} certain commodities such as Beef compared to Chickens had significantly more variance in terms of their range. For instance, Chickens production has a clear upward trend with slight variance over time. In contrast, the Beef production range significantly varies with abrupt changes year to year. Therefore, this is a factor that made a considerable difference in the linear fit $R^2$ score for the baselines. For the Linear Regression with Polynomial Features model, Beef’s $R^2$ score value was -0.1384 whereas Chickens was 0.5373 which had a much better linear fit for the data supporting our conclusion about data variance.

\subsection{DeepAg Outcomes}
\subsubsection{Forecasting Commodities' Production}
We employed the multi-step multivariate time series forecasting technique to accurately predict the commodity's production in the future. Multivariate time-series forecasting enables us to predict a future dependent variable $y$, based on more than one independent variable $x$. Here, our dependent variable $y$ is the commodity production, and independent variables $x$ are the historical production data, highest causation, highest correlation index, and historical outliers. Lastly, multi-step forecasting tunes the model to predict a certain number of time-steps ahead instead of only predicting \textit{one} future value. Our model had a look-back of \textit{60} time-steps from the past to forecast approximately \textit{30} multi-steps into the future. Based on the commodity since every commodity has a different number of data points, our model was able to predict approximately \textit{five years} ahead.

Table \ref{tab:DeepAgResults} indicates the results obtained using our DeepAg approach. For each commodity production, there is a prediction value and an RMSE score. These are generated \textit{with} and \textit{without} the outliers model for a total of 4 data points per commodity. The prediction value is the last data point from the commodity forecasting and the \textit{RMSE} score is mathematically calculated using Equation \ref{eq:RMSEEq}. Each prediction is paired with an \textit{RMSE} value to measure error rate and also to evaluate against the baselines. For each commodity, the most relevant indices are noted as \textit{(Highest Causation, Highest Correlation)} and \textit{(Highest Causation)} if they are the same. It follows as Beef \textit{(DOW)}, Butter \textit{(Gold)}, Cheese \textit{(Gold)}, Chickens \textit{(DOW)}, Duck \textit{(S\&P 500)}, Eggs \textit{(DOW)}, IceCream \textit{(S\&P 500, DOW)}, Lamb and mutton \textit{(Gold, DOW)}, Milk \textit{(S\&P 500, Gold)}, Other poultry \textit{(Oil, VIX)}, Pork \textit{(DOW)}, Sherbet \textit{(S\&P 500, DOW)}, Turkeys \textit{(DOW)}, Veal \textit{(DOW)}, Water Ices \textit{(VIX, DOW)}.

\begin{table}[t]
\scriptsize{
\begin{tabular}{|p{0.15\linewidth}|r|r|r|r|}
\hline
\textbf{Commodity} &
  \multicolumn{1}{p{0.14\linewidth}|}{\textbf{RMSE With Outliers}} &
  \multicolumn{1}{p{0.14\linewidth}|}{\textbf{Prediction With Outliers}} &
  \multicolumn{1}{p{0.13\linewidth}|}{\textbf{RMSE W/O Outliers}} &
  \multicolumn{1}{p{0.14\linewidth}|}{\textbf{Prediction W/O Outliers}} \\ \hline
\textbf{Beef}            & 0.164          & 2060199936  & 0.12  &  1960800000 \\ \hline
\textbf{Butter}          & 0.001          & 164524992   & 0.009 & 164524992                         \\ \hline
\textbf{Cheese}          & 0.004          & 991345984   & 0.058 & 991345984                         \\ \hline
\textbf{Chickens}        & 0.258          & 3531831040  & 0.197 & 3302286080                        \\ \hline
\textbf{Duck}            & 0.001 & 6348000     & 0.001 & 11891000                          \\ \hline
\textbf{Eggs}            & 0.708          & 8599600128  & 0.501 & 8599600128                        \\ \hline
\textbf{IceCream}        & 0              & 12010000    & 0.004 & 82232000                          \\ \hline
\textbf{Lamb \newline and mutton} & 0.001 & 12000000    & 0.001 & 10100000                          \\ \hline
\textbf{Milk}            & 1              & 18872999936 & 1     & 16984999936                       \\ \hline
\textbf{Other \newline poultry}   & 0   & 142000      & 0     & 100000                            \\ \hline
\textbf{Pork}            & 0.158 & 2022099968  & 0.118 & 2157299968                        \\ \hline
\textbf{Sherbet}         & 0     & 3541000     & 0     & 3541000                           \\ \hline
\textbf{Turkeys}         & 0.024          & 523214016   & 0.028 & 458169984                         \\ \hline
\textbf{Veal}            & 0.001 & 6700000     & 0     & 6000000                           \\ \hline
\textbf{Water Ices}      & 0     & 4647000     & 0     & 4647000                           \\ \hline
\end{tabular}}
\caption{\label{tab:DeepAgResults}DeepAg LSTM model results}
\end{table}

\begin{figure}[h]
\centering
\includegraphics[width=\linewidth]{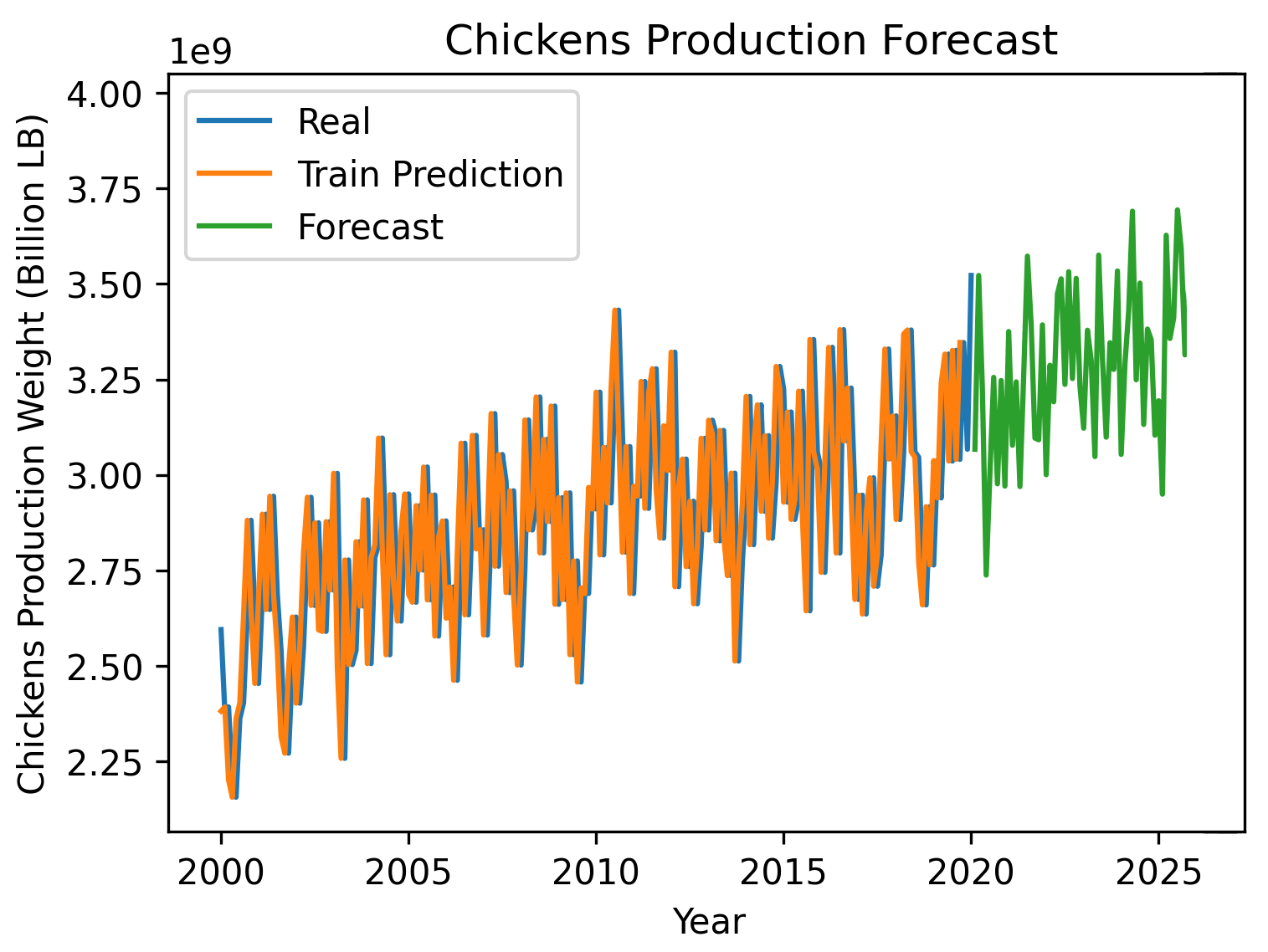}
\caption{Chickens production forecast 2020-2025}
\label{fig:chickensForecast}
\end{figure}

\begin{figure}[h]
\centering
\includegraphics[width=\linewidth]{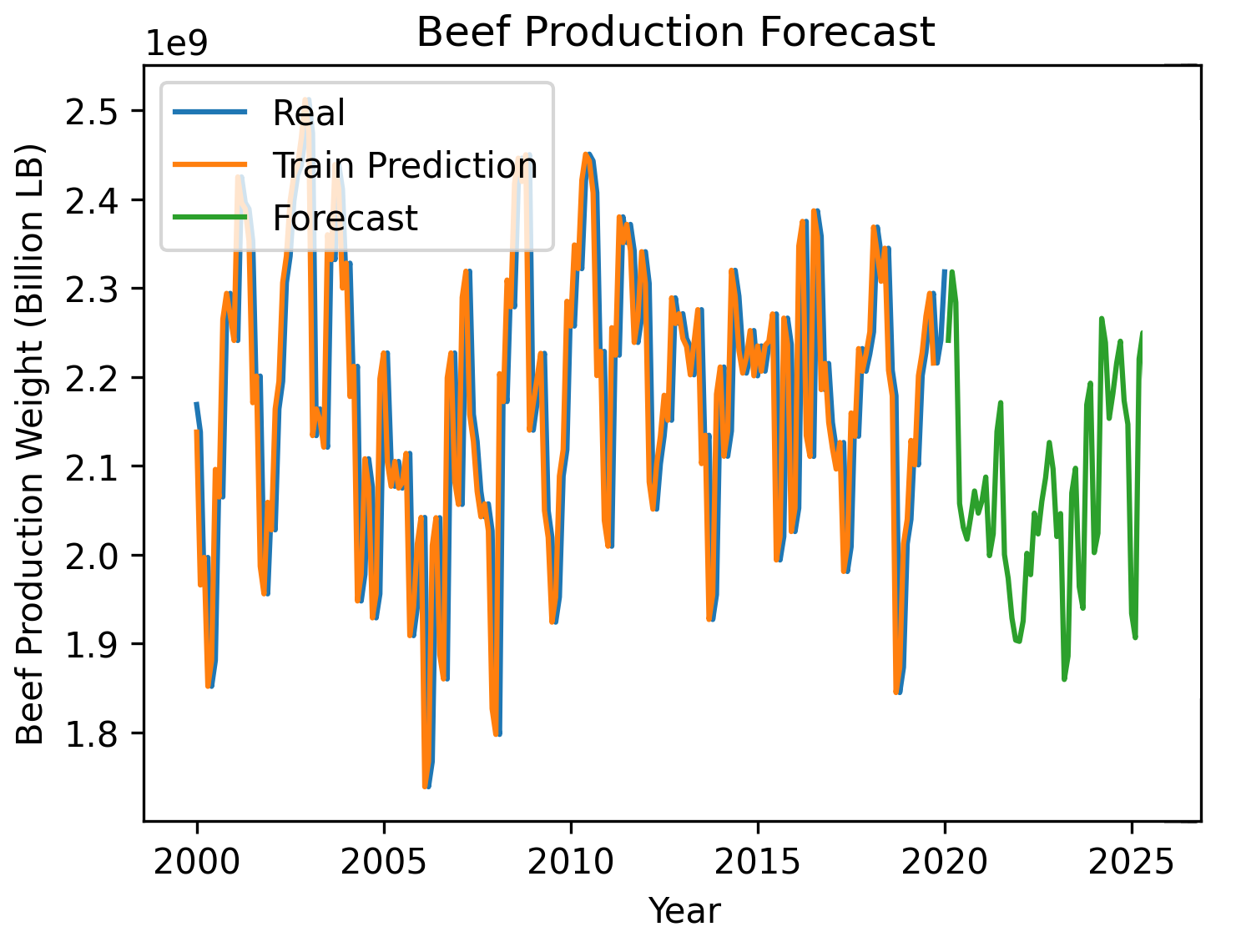}
\caption{Beef production forecast 2020-2025}
\label{fig:beefForecast}
\end{figure}

\subsection{The Effect of Outlier Detection}

The Figures \ref{fig:chickensForecast} and \ref{fig:beefForecast} illustrate the overall prediction fit and forecast for two of the commodities, Chickens and Beef. It’s evident that the model had a close train prediction compared to the real values and was able to accurately fit the data in addition to the forecasting. From Table \ref{tab:DeepAgResults}, the RMSE values demonstrate the high prediction accuracy of our DeepAg approach compared to the baselines. The lower the RMSE value, the better our approach is able to fit the production data and predict accurately. For instance, the RMSE for Beef production was 0.164, 0.120, 0.201 using DeepAg with outliers, DeepAg without outliers, and with Linear Regression with Polynomial Features baseline model, respectively, indicating that DeepAg with isolation forest outperforms DeepAg without outliers and the baselines. Out of the 15 commodities, 12 commodities had a lower RMSE with DeepAg - presented in Table \ref{tab:comparisonRMSE}. For the remaining three commodities, baseline RMSE was lower and this may be because those are consistent commodities (Chickens, Eggs, Milk) and tend to have a clear upward trend that's better predicted by linear regression models.

Table \ref{tab:comparisonRMSE} supports that outlier events are useful and contribute to a higher accuracy when predicting commodities' production. Outlier events are the cause of sudden upward or downward spikes in economics, and in our work, we have captured that ambiguity through a more formal process. The use of outliers as one of the input features allows the LSTM model to learn and optimize for these extreme events.

\begin{table}[]
\scriptsize{
\begin{tabular}{|p{0.15\linewidth}|p{0.24\linewidth}|p{0.20\linewidth}|p{0.20\linewidth}|}
\hline
\multicolumn{1}{|r|}{\textbf{Commodity}} &
  \textbf{Linear Regression with Polynomial Features RMSE} &
  \textbf{DeepAg With Isolation \newline Forest RMSE} &
  \textbf{DeepAg W/O Isolation \newline Forest RMSE} \\ \hline
\textbf{Beef}            & 0.201          & 0.164 & \textbf{0.120}          \\ \hline
\textbf{Butter}          & 0.138          & \textbf{0.001} & 0.009          \\ \hline
\textbf{Cheese}          & 0.139          & \textbf{0.004} & 0.058          \\ \hline
\textbf{Chickens}        & \textbf{0.131} & 0.258          & 0.197          \\ \hline
\textbf{Duck}            & 0.146          & \textbf{0.001} & \textbf{0.001} \\ \hline
\textbf{Eggs}            & \textbf{0.134} & 0.708          & 0.501          \\ \hline
\textbf{IceCream}   & 0.201          & \textbf{0.000} & 0.004          \\ \hline
\textbf{Lamb \newline and mutton} & 0.109          & \textbf{0.001} & \textbf{0.001} \\ \hline
\textbf{Milk}            & \textbf{0.093} & 1.000          & 1.000          \\ \hline
\textbf{Other \newline poultry}   & 0.224          & \textbf{0.000} & \textbf{0.000} \\ \hline
\textbf{Pork}            & 0.127          & 0.158          & \textbf{0.118} \\ \hline
\textbf{Sherbet}    & 0.192          & \textbf{0.000} & \textbf{0.000} \\ \hline
\textbf{Turkeys}         & 0.177          & \textbf{0.024} & 0.028          \\ \hline
\textbf{Veal}            & 0.137          & 0.001          & \textbf{0.000} \\ \hline
\textbf{Water Ices} & 0.287          & \textbf{0.000} & \textbf{0.000} \\ \hline
\end{tabular}}
\caption{\label{tab:comparisonRMSE}Comparison of best baseline model and DeepAg}
\end{table}

\section{Conclusions and Discussions}

In this manuscript, we proposed a novel approach called DeepAg that employs econometrics and leverages DL to measure the effect of outlier events when predicting the relationship between five commonplace financial indices and agricultural commodities. We compared DeepAg with existing techniques in terms of prediction accuracy. Our findings indicate that DeepAg's use of isolation forests enables it to outperform other techniques and offers a highly accurate prediction of the multivariate relationship in precision farming.

\subsection{Outlier Events}
Merging outlier events with production forecasts also reveals more accurate insights. A global supply shock for commodities, weather events, or an important international affair can affect production and can cause sudden spikes. Typically, these outlier events are sudden and cannot be planned for. During these times, producers are left with little insight on how their production will be affected. For instance, the COVID-19 pandemic resulted in a global supply shock for many essential commodities. The event created a large spike in the demand for products such as Beef and Chickens amongst other commodities and caused the price of these commodities to rise. This caused a shortage of essential commodities and caused the price of these commodities to rise even higher creating a cycle until the production supply of these commodities exceeded market demand. Our approach could potentially help producers plan for such events. Moreover, spikes in demand often contribute to global food waste where producers/retailers/consumers typically purchase more than their needs and often end up wasting the excess produce. The supply chain can be better equipped with future production trends with our approach and strategically release products in appropriate batches to mitigate such issues. We envision that these methods are also useful in cases of detecting cyberbiosecurity attacks on national infrastructure such as water management systems and supply chains.

\subsection{Public Policy Insights}
Predicting agricultural production is essential for feeding the world in the years ahead. The amount of agricultural production has a direct effect on supply, demand, and trade.  While we demonstrate our approach at the aggregate level, they can also be used at micro scales such as at a farm or county.  The aggregate analysis can particularly aid in shaping polices, since USDA, the Food and Agriculture Organization (FAO) and many other nations rely on country-specific as well as global forecasts to set policy parameters. Seasonal pattern data can also be used as a source of ground truth to determine trends and anticipate future demand. For example, note the prediction of chicken and beef production to 2025, which accounts for potential outlier events in the future; they are sharply different from existing straight-line forecasts and provide a bounded pathways for policy and other decisions. Our forecasting approach can help producers determine how low the production should drop and help them take preventative measures to continue operating even when production demand is low. During the winter, the producers may reduce the number of workers to save costs, minimize distribution, and reduce production volume. In summary, DeepAg can positively affect agriculture through on-time outcomes and can increase overall farm performance using Deep Learning.


\bibliographystyle{aaai} 
\bibliography{References}

\end{document}